\documentclass[letterpaper]{article} 
\usepackage{aaai21}  
\usepackage{times}  
\usepackage{helvet}  
\usepackage{courier}  
\usepackage[hyphens]{url}  
\usepackage{graphicx} 
\usepackage{natbib}  
\usepackage{caption} 
\usepackage{algorithm}
\usepackage{algorithmic}
\usepackage{comment}

\usepackage{newfloat}
\usepackage{listings}
\floatstyle{ruled}
\newfloat{listing}{tb}{lst}{}
\floatname{listing}{Listing}

\usepackage{adjustbox}
\usepackage{lipsum}  
\usepackage{tikz}
\usetikzlibrary{positioning}
\usetikzlibrary{backgrounds}
\usepackage{pgfplots}
\pgfplotsset{width=10cm,compat=1.9}
\usepgfplotslibrary{groupplots}
\usetikzlibrary{pgfplots.statistics}
\usepackage{booktabs}
\usepackage{tabularx}
\usepackage{caption}
\usepackage{subcaption}
\usepackage{multirow}
\usepackage{amsmath}
\usepackage{tikzscale}
\usepackage{amsfonts}

\usepackage{makecell}

\title{How Much User Context Do We Need? Privacy by Design in \\ Mental Health NLP Applications}
\author{
    Ramit Sawhney \textsuperscript{\rm 1} \textsuperscript{\rm 3},
    Atula Tejaswi Neerkaje \textsuperscript{\rm 1} \textsuperscript{\rm 4},
    Ivan Habernal \textsuperscript{\rm 2},
    Lucie Flek \textsuperscript{\rm 1}
    \\
}
\affiliations{

    \textsuperscript{\rm 1} Conversational AI and Social Analytics (CAISA) Lab, \\ Department of Mathematics and Computer Science, University of Marburg, Germany \\
    \textsuperscript{\rm 2} Trustworthy Human Language Technologies, \\ Department of Computer Science,  Technical University of Darmstadt, Germany\\
    \textsuperscript{\rm 3} Georgia Institute of Technology, USA \\
    \textsuperscript{\rm 4} Manipal Institute of Technology, India \\
    
}

\begin{document}
\maketitle
\begin{abstract}
Clinical NLP tasks such as mental health assessment from text, must take social constraints into account - the performance maximization must be constrained by the utmost importance of guaranteeing privacy of user data. Consumer protection regulations, such as GDPR, generally handle privacy by restricting data availability, such as requiring to limit user data to 'what is necessary' for a given purpose. In this work, we reason that providing stricter formal privacy guarantees, while increasing the volume of user data in the model, in most cases increases benefit for all parties involved, especially for the user. We demonstrate our arguments on two existing suicide risk assessment datasets of Twitter and Reddit posts. We present the first analysis juxtaposing user history length and differential privacy budgets and elaborate how modeling additional user context enables utility preservation while maintaining acceptable user privacy guarantees. 


\end{abstract}

\section{Introduction}
\label{section:introduction}


A growing body of work shows that NLP models can effectively support professional human counselors in identifying mental health risk markers in online user behavior, for example to aid suicide risk assessment~\citep{mccarthy2010internet, de2016discovering, 10.1001/jamapsychiatry.2020.1060,shing-etal-2018-expert}.
An early intervention by a counselor can be crucial, as 80\% of patients at suicidal risk do not undergo medical treatment. On the contrary, people turn to online platforms, with 8 out of 10 people disclosing their suicidal plans~\cite{golden2009truth, robinson2016social}. It is therefore desirable to train analytical models for these platforms, however, without unnecessarily exposing personal details of the patients in the training data.

State-of-the-art models for this task typically employ large contextual models \citep{matero2019suicide,losada2019overview,zirikly-etal-2019-clpsych}. Leveraging additional data, such as user's history on social media, augments the predictive power \citep{zirikly-etal-2019-clpsych, sawhney2021a}. 
However, as every machine learning model, these can be prone to learning undesired data artifacts, for example users mentioning certain locations, persons, or other rare pieces of information can be systematically misclassified as suicidal.

Moreover, privacy risks in such NLP models might lead to severe individual consequences. Not only can adversaries exploit sensitive artifacts from the models, but in extreme cases they can target vulnerable users~\citep{Hsin2016}. 

Although consumer data laws address protection of vulnerable users to some extent, e.g., by limiting the amount of necessary personal data for processing, data minimization does not ensure increased privacy protection. In contrary, this can be even counterproductive, as reducing user data may lead to less robust classifiers and more artifacts learned.
At the same time, advanced approaches such as \textit{differentially private learning}~\citep{dwork2014algorithmic} can be taken to abstain from using the privacy-breaching information in the learning process. However, these approaches typically lead to lower predictive power, making the classifiers less useful in practice~\citep{alvim2011differential}.

This raises an obvious empirical research question - \textbf{how much user data shall we use?} If data minimization does not increase privacy protection of users, can we include \emph{more} data instead and protect its privacy by design?
If so, which decisions do we need to consciously make in the model design? \textbf{What are the trade-offs between increasing formal user privacy preservation guarantees, maintaining predictive power, and minimizing users' training data?} Is there a way to find optimal thresholds for these?

\textbf{In this paper, we present the first empirical analysis juxtaposing user history length and differential privacy budgets, demonstrating how modeling additional user context can compensate the performance loss while keeping acceptable privacy guarantees.} 
Specifically, we demonstrate our approach on two suicide risk assessment datasets of Twitter and Reddit posts, as these tasks are vulnerable to user harm by privacy leaks \cite{o2019reviewing,rubanovich2022associations,mikal2016ethical}.
We qualitatively inspect data points where enforcing differential privacy leads to performance changes, quantitatively estimate the privacy leakage by adversarial attacks, and examine the impact of different levels of class imbalance. Based on our experiments, we argue that providing stricter formal privacy guarantees, while increasing the volume of user data in the model, in most cases increases the benefit for all parties involved, and that this effect shall be taken into account by future policy makers.



\section{Background and Prior Work}


A straight-forward approach to privacy protection in text is to identify sensitive user-revealing passages in a document and replace them with more general expressions \citep{hill2016effectiveness,alawad2020privacy}.  However, removing personally identifiable information is typically not sufficient, as the summary statistics of the dataset provide means to infer individual’s membership with high probability \citep{jones2007know,sweeney1997weaving}. Not only datasets, but also trained NLP models are vulnerable to attacks reconstructing the training data \citep{shokri2017membership}. 
 Privacy preservation efforts in NLP mostly follow adversarial approaches 
 to obscure sensitive user characteristics at training time, such that the representations learned are invariant to these attributes \citep{li2018towards,friedrich-etal-2019-adversarial,coavoux2018privacy}. 
 However, this approach offers only empirical privacy improvements, without any formal guarantees.

\begin{figure}[t]
    \centering
    \includegraphics[width=0.8\linewidth]{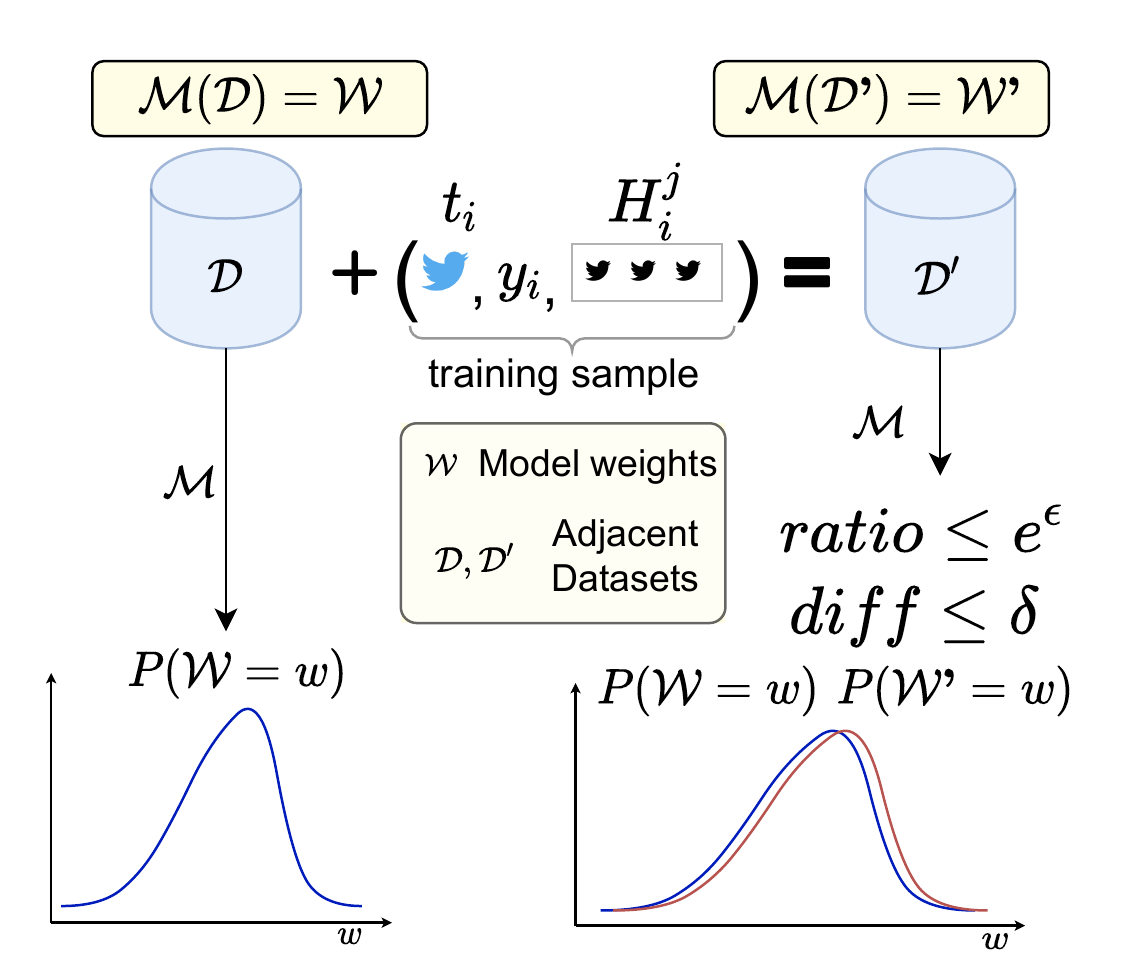}
    \caption{Essence of DP: With changes in a single training sample, the probability of obtaining a set of model weights remains approximately the same. Particularly, the ratio bounded by $e^{\varepsilon}$, and that difference bounded by $\delta$.}
    \label{fig:DiffPriv}
    \vskip -2ex
\end{figure}

\textbf{Differential Privacy (DP)} has been adopted as a standard framework to provide a quantifiable privacy guarantee \citep{dwork2014algorithmic}.
Intuitively, a randomized algorithm is differentially private if the output distributions from two neighboring input databases, that are identical in all but a single training example, are indistinguishable (Fig. \ref{fig:DiffPriv}). The basic idea is that the output of the mechanism is ``not much different" whether or not a single individual (example) is present in the dataset. The amount of ``not much different" is controlled by the parameter $\varepsilon$, the privacy budget, and has a probabilistic interpretation of how much privacy of a single individual can be ``lost" in the worst case.

In NLP, this paradigm is being mainly applied in the context of learning private word representations \citep{fernandes2018,beigi2019b,coavoux2018privacy} to prevent memorization in large language models~\citep{carlini2019secret,canfora2018nlp} rather than directly in the downstream NLP classifiers, which we explore in this paper. 

\textbf{DP and performance drop.} An increased privacy protection comes with a performance cost, typically leading to a decreased utility of the model~\citep{alvim2011differential,bassily2014private,shokri2015privacy,shokri2017membership}, mainly due to memorization being prevented \citep{van2018three}. While this effect has been described in machine learning research, its practical qualitative implications in the NLP field remain largely underexplored.

\textbf{DP and disparate impact.} Another challenge for applying privacy-preserving algorithms arises with data imbalance. Machine learning studies on other types of data (e.g. clinical records and images) show that the performance degradation is disparate, with minority subgroups of data suffering more utility loss \cite{shokri2017membership}.
Moreover, when stricter privacy guarantees are imposed, the utility gap widens disproportionally, i.e. the performance gap between the minority and majority subgroups is increasing \cite{bagdasaryan}. In other words, the less represented groups which already have lower accuracy end up losing more utility. This disparate impact of DP can occur even where classes are only slightly imbalanced, and isn't limited to strict privacy budgets \cite{farrand2020neither}. 

\textbf{DP and the choice of privacy budget.} The question of how to set the privacy budget ($\varepsilon$) has been present since the introduction of differential privacy and perceived as a ``social question''~\cite{dwork2006calibrating} with numerous mathematical answers proposed, such as utility-maximization optimizers~\cite{geng2014optimal} and interval estimations~\cite{naldi2015differential}. Voting mechanisms~\cite{kohli2018epsilon} gain popularity as they enable users to express their privacy budget; however, some studies find none to minimal links between privacy concerns reported by users and recorded behavior in a system with privacy warnings \citep{zimmerman2019investigating}. Some works argue for setting $\varepsilon$ based on inferred user traits instead \cite{vu2017personality}. In practice, values in the literature vary from as little as 0.01 to as much as 7, often with little to no justification. Recently, researchers examined DP settings of commercial companies \cite{tang2017privacy}, criticizing their generous $\varepsilon$ choice.
However, to date there is no clear consensus in the community on what strong or acceptable privacy budget values are. Generally, values under $1$ can be considered strongly private, as a well-known DP method called Randomized response ($\varepsilon = \ln(3))$ has been used for decades for highly sensitive questionnaires in social sciences \citep{warner1965}.

\textbf{DP versus data minimization.} Our motivation for contrasting the theoretical privacy preservation (as formalized by DP) with the data minimization approach originates from examining how the goal of personal data protection of the users is anchored in consumer data laws. For example the European Union's General Data Protection Regulation (GDPR) requires limiting user data \emph{to what is necessary concerning the purposes for which they are processed}, encouraging data minimization rather than privacy protection by design. The operationalization of these laws in machine learning applications, including NLP, has been controversial~\citep{tene2012big,biega2020}, with practical implications of various design choices often remaining rather opaque to the engineers and policy makers alike. 
Several data minimization studies have been conducted in the domain of large recommender systems, noting that the overall performance often does not drop after removing random data points \cite{chow2013differential} or drastically reducing user history length \cite{wen2018exploring}. However, the incurred changes disparately impacted result quality for specific users or user groups \cite{biega2020}. In this work, we conduct experiments to empirically examine such effects in the area of sensitive NLP tasks.

\section{Methodology}

\begin{figure*}
    \centering
    \includegraphics[width=0.9\textwidth]{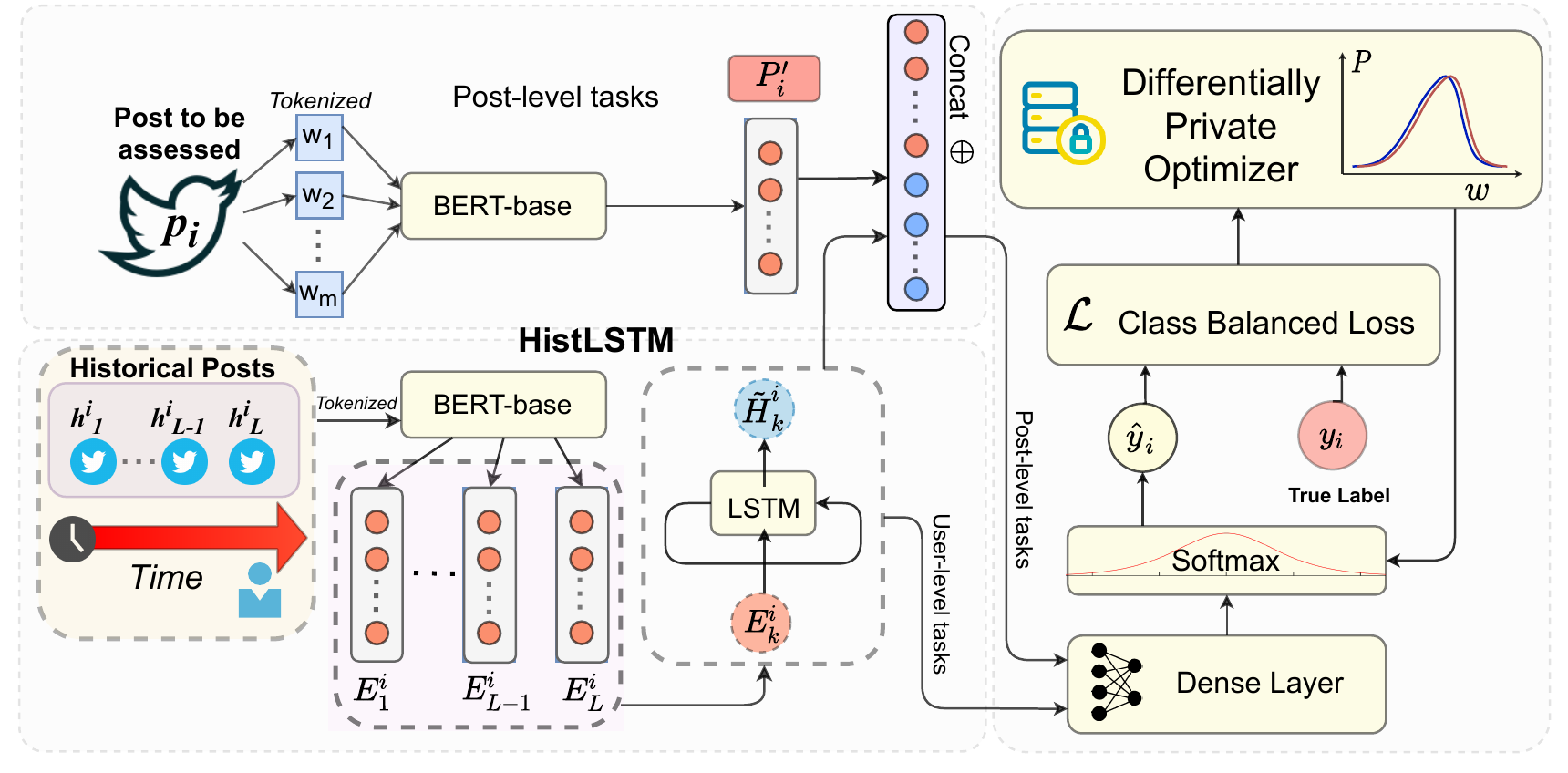}
    \caption{An overview of HistLSTM: We first encode all user posts. The historical post representations are sequentially modeled via an LSTM to obtain user historical context. For post-level tasks, it is concatenated with the post to be assessed and is used for classification. The model is optimized using a differentially private optimizer. }
    \label{fig:overview}
    \vskip -1ex
\end{figure*}
\label{section:methodology}

We explore to which extent varying amounts of users' historical data in combination with varying privacy budget impact performance of sensitive user-centric social NLP tasks, that is the suicide ideation and risk detection tasks.


Let $S_i$ be the $i$-th training sample associated with user $u_j$ from the set of $M$ users. Each sample consists of a sequence of historical posts authored by the particular user $u_j$ which we denote as $H_i^j=[h_1^i, h_2^i, \cdots, h_L^i]$ which occur strictly before the time step where the prediction is made. The task is a classification problem, namely to predict label $y_i$.

We study two kinds of user-centric classification tasks. First, for \emph{post-level} tasks, the suicidality label is associated with the most recent post to be assessed $p_i$. Here the user's historical posts provide an additional context. The training sample $S_i$ is hence the tuple $(p_i,y_i,H_i^j)$. Second, for \emph{user-level} tasks, the suicidality label is associated with the user $u_j$, with the sample $S_i$ representing the tuple $(y_i,H_i^j)$.

\subsection{Tasks and Data Studied}
We identify suicide ideation and risk detection on social media as prediction tasks which benefit from additional contextual data \citep{zirikly-etal-2019-clpsych, sawhney2021a}, with the need to protect private and sensitive data \cite{huckvale2019assessment,rubanovich2022associations}.
\paragraph{Suicide ideation (SI)}
A post-level binary classification task, for which we use an existing Twitter dataset~\citep{mishra-etal-2019-snap} comprising Twitter timelines of 32,558 unique users, spanning over ten years of historical tweets from 2009 to 2019, summing up to 2.3M unlabeled tweets. These tweets were identified using a lexicon of 143 suicidal phrases. Out of these, 34,306 latest tweets were labeled as $y_i \in $ \{\textit{Suicidal Intent Absent}, \textit{Suicidal Intent Present}\}. The dataset contains 3,984 suicidal tweets, indicating a high class imbalance of 9:1\footnote{The imbalance is much greater in the real world}.
\paragraph{Suicide risk -- Reddit (SR)}
A user-level multi-class classification task, for which we use an existing dataset released by \citet{gaur2019knowledge}. The dataset consists of 9,127 Reddit posts by 500 users. Each user is associated with a suicide risk severity label $y_i \in$ \{\textit{Indication}, \textit{Ideation}, \textit{Behaviour}, \textit{Attempt}\} in increasing order of suicide risk. The average number of posts made by a user is 18.25 $\pm$ 27.45. 
\footnote{We provide further insights on the individual class distributions in the appendix.}.

\subsection{User-Contextual NLP Modeling}
We describe our base user-contextual model HistLSTM, which sequentially models user's historical posts. For post-level tasks, features obtained by HistLSTM are combined with the post to be assessed before the classification step.

\paragraph{Encoding posts}
We use the 768-dimensional representation of the \texttt{[CLS]} token obtained from BERT 
which tend to yield holistic representations of text, including posts on social media. We encode $\mathbf{P_i'} \in \mathbb{R}^{768}$ for a post to be assessed $p_i$ as $\mathbf{P_i'} = \text{BERT}_(p_i)$. Similarly, we encode each historical post $h_k^i$ to yield the representation $\mathbf{E_k^i} \in \mathbb{R}^{768} = \text{BERT}(h_k^i)$.

\paragraph{Historical context}
The sequential nature of historical posts makes LSTMs a natural method for contextual modeling.
Each historical post embedding $\mathbf{E_k^i} \in \mathbf{E^i}$ is fed sequentially to the LSTM to obtain the final hidden state $\mathbf{\widetilde{H}^i_L} \in \mathbb{R}^d = \text{LSTM}(\mathbf{E^i})$. For post-level tasks, HistLSTM jointly learns from the language of the post to be assessed along with the sequentially modeled representation of the user's timeline. For such tasks we apply the concatenation operation $\oplus$ to $\mathbf{P_i'}$ and  $\mathbf{\widetilde{H}^i_L}$ before feeding it to a dense layer with $\operatorname{ReLU}$ activation, followed by softmax. 
$$
    \begin{aligned}
          & \mathbf{\widehat{y}_i} = \operatorname{softmax}(\operatorname{ReLU}(W_y  (\mathbf{P_i'} \oplus \mathbf{\widetilde{H}^i_L}) + b_y)) \\
          \end{aligned}
$$
For user-level tasks, $\mathbf{\widetilde{H}^i_L}$ is directly fed to dense layer:
$$
    \begin{aligned}
          & \mathbf{\widehat{y}_i} = \operatorname{softmax}(\operatorname{ReLU}(W_y  (\mathbf{\widetilde{H}^i_L}) + b_y)) \\
          \end{aligned}
$$
 where $\mathbf{\widehat{y}_i}$ is the class probabilities vector and \{$W_y,b_y$\} are trainable network parameters.

\subsection{Differentially Private Optimization}

Differential privacy has been adopted as a standard framework for privacy-preserving analysis. DP protects privacy of each single individual in the dataset by introducing random noise into a mechanism that queries the dataset. When a trained model is differentially private, the $\varepsilon$ parameter limits the probability of an adversary to attack the model and reveal sensitive information from its training dataset. 
Formally, a mechanism $\mathcal{M}$  with domain $\mathcal{D}$ and range $\mathcal{R}$ represented by $\mathcal{M} : \mathcal{D} \rightarrow \mathcal{R}$ satisfies ($\varepsilon,\delta$)-differential privacy if for two adjacent datasets $d,d' \in \mathcal{D}$ that differ in one example and for any subset of outputs $\mathcal{S} \subseteq \mathcal{R}$ it holds that:
\begin{equation}
    \Pr[\mathcal{M}(d) \in \mathcal{S}] \leq \exp(\varepsilon) \cdot \Pr[\mathcal{M}(d') \in \mathcal{S}] + \delta
\end{equation}
Lower $\varepsilon$ implies better privacy. The term $\delta$, which is typically `cryptographicaly small,' is a relaxation of the `pure' $(\varepsilon, 0)$-DP. It enables much better composition of several private algorithms in exchange for a small probability of privacy `failure' \citep{abadi2016deep}.
In deep learning, the mechanism  $\mathcal{M} : \mathcal{D} \rightarrow \mathcal{R}$ can be considered as a training procedure $\mathcal{M}$ on a dataset $\mathcal{D}$ that returns a model in the space $\mathcal{R}$ \citep{bagdasaryan}.

Intuitively,  Differential Privacy requires that the probability of learning any particular set of model parameters stays roughly the same if we change a single training example in the training set Figure~\ref{fig:DiffPriv}.
We implement differential privacy based on the widely adopted \cite{ha2019differential} DP-SGD algorithm introduced by \citet{abadi2016deep}.

The key aspects of differentially private SGD are (1) clipping the per-batch gradient where its norm exceeds the clipping bound $C$ and (2) adding zero-centered Gaussian noise $\mathcal{N}$ parametrized by $\sigma$ to the aggregated per-sample gradients. Since the gradient computation is a function applied to the dataset and its output is privatized with DP, the resulting trained model will also be differentially-private. Intuitively, this means that by limiting the norm of the gradients and adding additional noise, we prevent the model from learning more than $\varepsilon$ amount of information from every particular training sample, regardless of how distinct it is from the others. However, for DP-SGD, an important issue is quantifying the maximum privacy loss of each training sample i.e computing the value of the privacy budget ($\varepsilon$).

We track the privacy budget ($\varepsilon$) using R\'enyi differential privacy \citep{mironov2017renyi}, a relaxation of $(\varepsilon,\delta)$-DP, which makes tracking the privacy budget easier. The moments accountant in \citet{abadi2016deep} can be considered as a special case of R\'enyi differential privacy. For a given clipping bound $C$ and Gaussian noise $\mathcal{N}$, the moments accountant ``tracks'' the privacy cost at each access to the training data, and accumulates this cost as the training progresses, eventually giving us the a final estimate of the privacy budget when training is stopped. We do not delve into those details here, but we point the reader to the appendix, and \citet{abadi2016deep} for extensive mathematical derivations of DP-SGD and the moments accountant.

\subsection{Network Optimization and Classification}

Often, in user-centric social NLP tasks, the posts of interest form only a small percentage of the data, leading to class imbalance. Thus, we apply Class-Balanced loss \citep{cui2019class} to train HistLSTM, which introduces a weighting factor that is inversely proportional to the number of samples per class, yielding the loss $\mathcal{L}$ as:
\begin{equation}
    \mathcal{L}(\mathbf{\widehat{y}_i}, y_i) = \text{CB}_{focal}(\mathbf{\widehat{y}_i}, y_i; \beta, \gamma)
\end{equation}
where $\text{CB}_{focal}$ is class-balanced focal loss, $y_i$ is the true label, and $\beta$ and $\gamma$ are hyperparameters.
 
We use a differentially private Adam optimizer, a variant of DP-SGD. This optimization procedure leverages gradient clipping and noise addition to calculate the model parameter update $\mathbf{\Delta{w}}$ in the loss landscape. The moments accountant simultaneously computes the privacy budget $\varepsilon$ at the end of each epoch.

\begin{table*}[h!]
\begin{tabular}{llrrrrrr} \toprule
& & \multicolumn{3}{c}{Suicide Ideation} & \multicolumn{3}{c}{Suicide Risk}  \\
\cmidrule{3-5} \cmidrule(lr){6-8}
Model & Architecture and Settings & M-F1 & R\textsubscript{s} & P\textsubscript{s} & F1 & Gr.\ R  & Gr.\ P \\ \midrule
\makecell[l]{CurrentPostRF \\ \citep{sawhney2018computational}}
&
\makecell[l]{Language features: LIWC, POS tags, \\ n-grams fed to a Random Forest classifier. \\ Only uses most recent post.}    & 0.54    & 0.51 & 0.47  & 0.47 & 0.49 & 0.45  \vspace*{0.3em} \\ 

\begin{tabular}[c]{@{}l@{}}CurrentPostLSTM\\\citep{coppersmith2018natural}\end{tabular}               &
\makecell[l]{Single-post model which utilizes GloVe \\embeddings of the post fed to an LSTM. \\ Only uses most recent post.} & 0.58  & 0.59 & 0.52 & 0.51 & 0.51 & 0.53 \vspace*{0.3em} \\
\makecell[l]{HistCNN \\ \citep{gaur2019knowledge}}
&
\makecell[l]{Concatenated post embeddings of all posts \\ fed to a CNN}  & 0.71* & {0.59}* & \textbf{0.72}\dag & 0.55* & 0.51 & \textbf{0.69}\dag \vspace*{0.3em} \\
\makecell[l]{HistDecay\\\citep{mathur2020utilizing}}
&
\makecell[l]{Ensemble of BiLSTM+Attention with \\ exponentially weighted embeddings \\ of historical posts} & \textit{0.73}* & \textbf{0.76}* &  0.58*  & \textit{0.56}* & \textit{0.60}*  & 0.57* \vspace*{0.3em} \\ \midrule 
HistLSTM (Ours)
& Our base model with no DP optimization & \textbf{0.75}\dag & \textit{0.71}* & \textit{0.65}\dag   &\textbf{0.59}\dag & \textbf{0.61}\dag  & \textit{0.62}\dag \vspace*{0.3em} \\

\begin{tabular}[c]{@{}l@{}}HistLSTM + DP (Ours)\end{tabular}
&
\makecell[l]{HistLSTM optimized with a strict privacy \\ budget ($\varepsilon=0.74$)}
& 0.68 & 0.61* &  0.53*  & \textit{0.56}\dag & 0.58* & 0.57* \\ \bottomrule
\end{tabular}
\caption{Mean of results obtained over 15 independent runs. * and $\dag$ indicate that the result is significantly ($p<0.005$) better than CurrentPostLSTM (Non-contextual model), and HistDecay (SOTA) under Wilcoxon’s Signed Rank test respectively. \textbf{Bold} and \textit{italics }indicates best and second best performance respectively.}
\label{table:perfcmp}
\end{table*}

\begin{figure}[t]
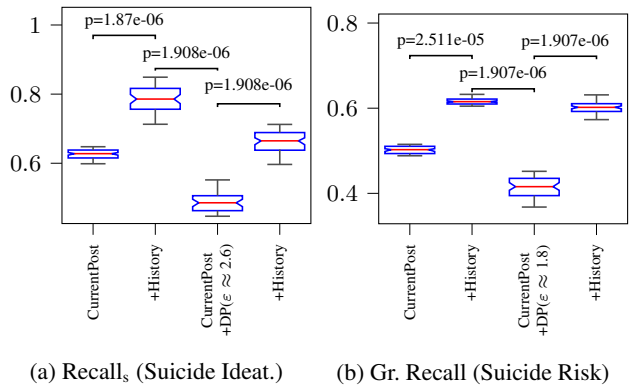

    \begin{subfigure}{0.49\linewidth}
    \includegraphics[width=\linewidth, height=4.5cm]{plots/recallbox.tikz}
    \caption{Recall\textsubscript{s} (Suicide Ideat.)}
    \end{subfigure}
    \begin{subfigure}{0.49\linewidth}
    \includegraphics[width=\linewidth, height=4.5cm]{plots/Grecall.tikz}
    \caption{Gr. Recall (Suicide Risk)}
    \end{subfigure}
    \caption{Confidence intervals for recall for the positive class over 10 independent runs for both tasks. ($p$) indicates the p-value under Wilcoxon's Signed Rank test. }
     \label{fig:boxplot}
\vskip -1ex
\end{figure}

\subsection{Evaluation Metrics}
We choose the evaluation metrics following existing work on suicide ideation detection \cite{gaur2019knowledge}. For the Suicide Ideation dataset, we use Recall on the minority class (\textit{Suicidal Intent Present}) as it is important to not misclassify posts with suicidal intent. We also report Macro F1 score for overall performance. For the Suicide Risk dataset, we use Graded Recall, Graded Precision and FScore, which are calculated using altered definitions of Recall and Precision to account for ordinal nature of risk labels~\cite{gaur2019knowledge}.

\subsection{Experimental Setup}
We split the twitter data using a temporal 70:10:20 (train:val:test) split, and the suicide risk dataset using a stratified 70:10:20 split. We use Macro F1/FScore on the validation set to select hyperparameters. Using grid search we explore: Hidden dimension $H^d$, Dropout $d$, $\beta$ and $\gamma$, learning rate $\eta$, batch size $N$. We set $\delta$ as a value smaller than the inverse of the size of the dataset \citep{abadi2016deep}. We manually set noise scale $\sigma$ and clipping bound $C$. It is important to note that DP-SGD is not parameterized by the desired privacy budget $\varepsilon$, but rather by the noise scale $\sigma$ and the clipping bound $C$, from which the final consumed privacy budget is computed. The general rule is that higher noise and lower clipping bounds yield tighter privacy budgets i.e lower values of $\varepsilon$.

\section{Results}
\label{section:results}

\subsection{What is the impact of User History Length?}
\label{subsec:comparison}

In line with previous work \citep{khattri2015your},
we observe that adding historical context of a user timeline leads to significant $(p<0.005)$ improvement in performance for both tasks, as illustrated in Figure \ref{fig:boxplot}. With longer history, there is consistent increase in performance, as shown in Figure \ref{fig:datavolume}.

We additionaly compare our model (\texttt{HistLSTM}) with other works that use single post only (\texttt{CurrentPost}), and those utilizing historical user context (\texttt{Hist}), in Table \ref{table:perfcmp}.
As expected, user-contextual models provide significant improvements compared to single-post models \citep{dadvar2013improving}. Among models with user history, temporally sequential models (\texttt{HistDecay}, \texttt{HistLSTM}) outperform bag-of-posts (\texttt{HistCNN}), likely due to their ability to capture the timeline build-up \citep{braadvik2008long}.

\subsection{What is the Impact of Privacy Budget Choice?}

\begin{figure}[t!]
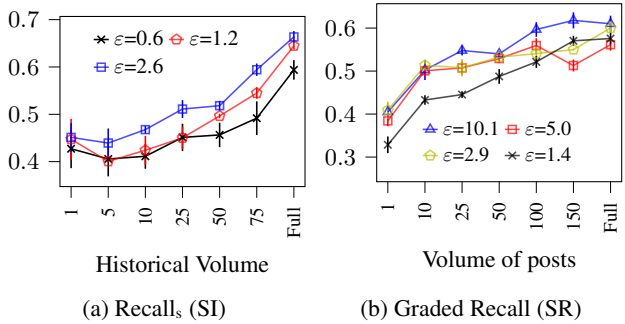

    \begin{subfigure}{0.49\linewidth}
    \includegraphics[width=\linewidth, height=3.7cm]{plots/volumeRecallSuicide.tikz}
    \caption{Recall\textsubscript{s} (SI)}
    \label{fig:hatef1}
    \end{subfigure}
    \begin{subfigure}{0.49\linewidth}
    \includegraphics[width=\linewidth, height=3.7cm]{plots/GrecallPlot.tikz}
    \caption{Graded Recall (SR)}
    \label{fig:haterecall}
    \end{subfigure}
    \caption{Performance (Recall) variations with increasing historical volume (number of historical tweets) and privacy budgets $(\varepsilon)$ over 10 independent runs.}
        \label{fig:datavolume}
        \vskip -1ex

\end{figure}

Incorporating differential privacy leads to performance drops, as shown in Figure \ref{fig:boxplot}.
With less privacy (higher $\varepsilon$), the performance increases for a fixed window of historical volume (Figure \ref{fig:datavolume}).
However, stricter private models ($\varepsilon<$1) with historical user context (\texttt{HistLSTM} + \texttt{Strict DP}) outperform non-private single-post methods, as shown in Table \ref{table:perfcmp}. For the SR dataset, strictly private models are on par with contextual models like \texttt{HistCNN}.
Interestingly, models with a relaxed privacy budget perform on par with SOTA user-contextual methods, showing although there is a performance-privacy trade-off, there exist optimal balances where both aspects can be equitable.

\subsection{How Vulnerable to Privacy Attacks are the Models?}
To aid interpretability of the privacy budget and trade-offs involved, we present empirical results from conducting black-box membership inference attacks \cite{shokri2017membership} for various values of the privacy budget as shown in Table \ref{table:privcmp}. We evaluate adversarial robustness using the privacy leakage metric \cite{jayaraman2019evaluating}, which is defined the difference between the true positive (TP) and false positive (FP) rate of the adversary. The key takeaways from our experiments are that data minimization alone is not sufficient to preserve privacy, reinforced by the observation that the non-private single-post model (\texttt{CurrentPost}) shows significant privacy leakage. The non-private HistLSTM performs best, at the cost of the highest privacy leakage. The incorporation of DP drastically reduces the adversarial efficacy with decreasing privacy budgets ($\varepsilon$), however at the cost of high performance degradation of the single-post model. Contrastingly, the addition of historical context to differentially private models increases performance while maintaining adversarial robustness. Specifically, even for strict privacy budgets ($\varepsilon=0.6$), HistLSTM shows significant utility,  maintaining similar adversarial robustness as single-post models.

\subsection{Can We Compensate stricter Privacy Budget by richer User History?}
We observe adding historical context to the DP model leads to significant performance improvements $(p<0.005)$ with moderate DP budgets ($\varepsilon$ $\approx$2.6, $\varepsilon$ $\approx$1.8) (Figure \ref{fig:boxplot}). We note that DP models accounting for historical context outperform the non-contextual, non-private models in both tasks.
Such models with user context and relaxed privacy budgets are able to match models with historical context (\texttt{HistDecay}), while models with strict privacy budgets ($\varepsilon<1$) outperform single-post methods (\texttt{Current}), as shown in Table \ref{table:perfcmp}.

\begin{table}[]
\begin{adjustbox}{width=\linewidth,keepaspectratio}
\begin{tabular}{@{}lrrrrrr@{}}
\toprule
\multirow{2}{*}{\textbf{Model}} & \multicolumn{2}{c}{\textbf{No DP}}                                                                                              & \multicolumn{2}{c}{\textbf{$\varepsilon$ $\approx$2.6}}                                                                                            & \multicolumn{2}{c}{\textbf{$\varepsilon$ $\approx$0.6}}                                                                                            \\ \cmidrule(lr){2-3} \cmidrule(lr){4-5} \cmidrule(lr){6-7} 
                                & \multicolumn{1}{l}{\textbf{\begin{tabular}[c]{@{}l@{}}PL\end{tabular}}}$\downarrow$ & \multicolumn{1}{l}{\textbf{M. F1}}$\uparrow$ & \multicolumn{1}{l}{\textbf{\begin{tabular}[c]{@{}l@{}}PL\end{tabular}}}$\downarrow$ & \multicolumn{1}{l}{\textbf{M. F1}}$\uparrow$ & \multicolumn{1}{l}{\textbf{\begin{tabular}[c]{@{}l@{}}PL\end{tabular}}}$\downarrow$ & \multicolumn{1}{l}{\textbf{M. F1}}$\uparrow$ \\ \midrule
CurrentPost                     & 0.19                                                                                    & 0.54                                  & 0.03                                                                                    & 0.45                                  & 0.01                                                                                    & 0.43                                  \\
HistLSTM                        & 0.38                                                                                    & 0.75                                 & 0.05                                                                                    & 0.69                                  & 0.02                                                                                    & 0.62                                  \\ \bottomrule
\end{tabular}
\end{adjustbox}
\caption{Model performance on the SI dataset with corresponding privacy leakage (PL) on conducting membership inference attacks for various values of privacy budget ($\varepsilon$).}
\label{table:privcmp}
\end{table}

\begin{figure}[t!]
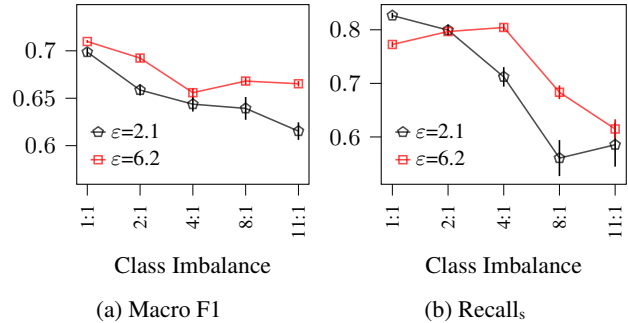

    \begin{subfigure}{0.49\linewidth}
    \includegraphics[width=\linewidth, height=3.7cm]{plots/imbalanceF1.tikz}
    \caption{Macro F1}
    \label{fig:f1Imbal}
    \end{subfigure}
    \begin{subfigure}{0.49\linewidth}
    \includegraphics[width=\linewidth, height=3.7cm]{plots/imbalanceRecall.tikz}
    \caption{Recall\textsubscript{s}}
    \label{fig:recallImbal}
    \end{subfigure}
    \caption{Changes in performance metrics with increasing class imbalance over 10 different runs (SI).}
        \label{fig:ClassImbalance}
        \vskip -1ex

\end{figure}

We now contrast how performance varies over different user contexts (amount of historical volume) and privacy budgets in Figure~\ref{fig:datavolume}. Specifically, we focus on Recall for the minority class (\textit{Suicidal Intent Present}), and Graded Recall (SR dataset) as the metrics of interest (Figure \ref{fig:datavolume}). Also, for DP models, the performance for both tasks consistently increases with increasing historical volume. For more relaxed privacy budgets ($\varepsilon>2$), performance with full post history (over 50 tweets) is comparable to the non-private models. For lower privacy budgets the performance suffers significantly, yet surpassing non-private, single-post (\texttt{Current}) models (Table \ref{table:perfcmp}). The tradeoff between privacy and utility 
coupled with the historical data volume, is somewhat task-dependent.
For the suicide ideation task, a more distant history (50 tweets) contributes to notable utility compensation (outperforming non-private counterparts with no historical context), even with strict privacy budgets, while the suicide risk model saturates earlier.\footnote{We observe very similar patterns for other metrics, and using time-based history cut-offs (as opposed to volume-based cut-offs). Results are provided in the appendix.}

\begin{figure*}[t]
    \centering
    \includegraphics[width=\textwidth]{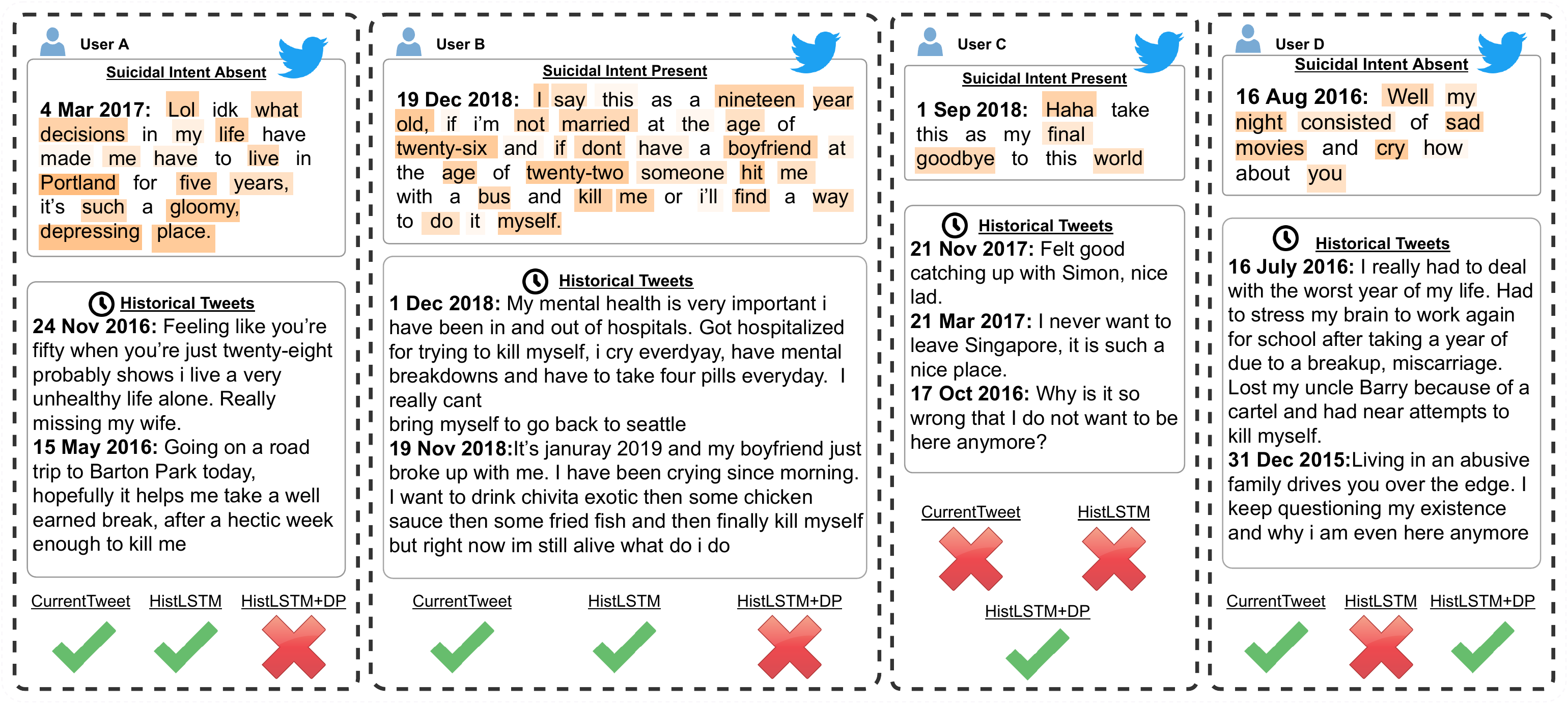}
    \caption{We study tweets of four users (SI dataset) and how a non-contextual model (CurrentPost) performs in comparison to a contextual model (HistLSTM, complete user history) and HistLSTM with strict ($\varepsilon=2.6$) privacy budget. We also visualize token self-attention of HistLSTM for descriptiveness (more intense means higher attention weight).}
    \vskip -2ex
    \label{fig:qa}
\end{figure*}
\label{subsec:classimbalance}

\subsection{Probing the Impact of Class Imbalance}
To study the effect of class imbalance on model performance and privacy, we train the model with varying class balances in Figure \ref{fig:ClassImbalance}. For these experiments, we randomly sub-sample the full dataset before performing the split. We perform the random sub-sampling for each run of the experiments to minimize any sampling biases. We observe that for a given privacy budget ($\varepsilon$), the Macro F1 gradually decreases with increasing class imbalance. The recall of DP models drops much faster for the minority class when higher privacy guarantees are required, with drastic drops for the strict budgets ($\varepsilon<2.6)$. Our observations align with existing work which suggest that privacy and fairness are fundamentally at odds under empirical risk minimization \citep{chang2021privacy,agarwal2021trade, tran2021differentially}. Note that the performance degradation is disparate, with the underrepresented user groups (those at high suicide risk) suffering more utility loss. 
Interestingly, at the most extreme imbalance settings, we note slight performance improvements. Possibly the tighter privacy guarantees degrade the utility on minority class so much that the model becomes more random, sometimes leading to an improvement \cite{farrand2020neither}.

\subsection{Qualitative Analysis}
To add further insight beyond the challenging mathematical formulations of interpreting DP
, we present a qualitative insight 
in Figure \ref{fig:qa}. 
Generally, we observe that augmenting models with DP leads to a failure in capturing relationships between the class labels and specific personal data related to named entities and numbers, which is sometimes desirable. 
For example, we see that tweets of User A and User B contain personal details such as locations and age. The non-private models likely leverage and memorize such details~\citep{carlini2019secret}
, whereas the DP-based model misclassifies the tweet.
For user C, models are thrown off by the positive words. Due to most of the historical tweets containing personal data, the tweet from 2016 may seem of higher importance to the DP variant, which correctly predicts the tweet to be suicidal. 




\subsection{Choice of Privacy Budget}


In our experiments, we observe moderate privacy budgets ($0.5 < \varepsilon < 4.5$) to notably decrease the sensitivity to names and digits (based on our subjective error analysis) while yielding acceptable performance when combined with full user timelines. Models with more benevolent (higher) $\varepsilon$ choices tend to perform similarly to non-private settings.


\section{Discussion}
\label{section:discussion}

\subsection{Trade-offs: Privacy, Performance, Data Volume}

Overall, settings where performance loss caused by high privacy guarantees is compensated by high data volumes, appear appealing and socially desirable.
However, there might be cases in practice, where this choice is suboptimal, e.g. when highly imbalanced classes can result to socially and economically problematic disparate impact 
\cite{bagdasaryan}.
Furthermore, experiments of \citet{shokri2017membership} also indicate disparate privacy vulnerability - smaller groups are more vulnerable to privacy disclosure under DP, which is in case of suicidal behavior a considerable design issue.
Overall, an intervention in the form of stricter privacy budget (lower $\varepsilon$) can increase the risk of high outcome bias (error disparity), while a stricter data minimization may increase the risk of high input bias (selection bias or over-amplification)~\cite{khani2020removing}.

Data minimization can be also consciously applied for purposes beyond privacy preservation, e.g. \citet{vincent2019data} explore the ``data strike'' as a form of collective action of users against technology platforms, starving business-critical models of training data. The authors find the impact of data strikes rather limited. Moreover, the resulting data voids can be exploited by malicious users to bias the system \citep{golebiewski2018data}.
Sometimes overall data minimization can be a prevalent goal for environmental reasons, as large models leave a major carbon footprint \citep{strubell-etal-2019-energy}.

\newpage
\section{Conclusions}
\label{section:conclusion}

In this paper we have shown that leveraging \emph{more} user-centric data enables competitive utility in prediction tasks and, at the same time, protects individual's privacy, provided the training procedure uses differential privacy with a strict privacy budget.
This setup increases the benefit for all parties involved, especially in sensitive tasks such as mental health assessment, where a strong privacy protection, hand in hand with low error rate, has been the most relevant factor for the patients themselves \cite{aledavood2017data,Hsin2016,torous2016psychiatry}. Strong privacy guarantees also enable sharing the research outcomes, such as deploying the trained models, in this sensitive domain~\cite{lehman2021does}.

We present the first analysis juxtaposing user history length and differential privacy budgets. We demonstrate on two publicly available (Twitter and Reddit) datasets for suicide risk assessment how increasing the historical context used for each user helps compensate for the performance loss of DP models even with strict privacy budgets ($\varepsilon \approx 0.6$), i.e., preserving high formal privacy guarantees. 

We qualitatively inspect data points where enforcing differential privacy leads to performance drops and surprising performance improvements. We observe decreased importance of named entities and numerals in the model, such as the author's age and location, potentially leading to preferable, robust representations. 

We show that increasing the imbalance of the predicted classes impacts the utility of DP models, given the disparate degradation in performance for minority classes, i.e., tweets with suicide intent. We also observe that a stricter choice of privacy budget results in more pronounced disparate degradation, i.e. steeper drop in recall for the minority class.

While we generally endorse applying private and data-minimizing model designs on user-centric NLP tasks, we point out that the disparate impact of such choice shall be reflected in design decisions.





\section*{Broader Impact \& Ethical Considerations}
\label{sec:ethics}



Emphasizing the sensitive nature of this work, we acknowledge the trade-off between privacy and effectiveness.
To avoid coercion and intrusive treatment, we work within the purview of acceptable privacy practices suggested by \citet{chancellor}. 
We paraphrase all examples shown in this work using the moderate disguise scheme \cite{bruckman2002studying} to protect user privacy~\citep{chancellor}. 
 We utilize publicly available data in a purely observational~\citep{DBLP:journals/corr/NorvalH17}
, and non-intrusive manner. For all tasks, we specifically only use existing datasets, and all user data is kept separately on protected servers linked to the raw text and network data only through anonymous IDs.

We acknowledge that suicidality is subjective, the interpretation of this analysis may vary across individuals on social media~\citep{puschmannbad}, and we do not know the true intentions of the user behind the post. Care should be taken so as to not to create stigma, and interventions must hence be carefully planned by consulting relevant stakeholders, such as clinicians, designers, and researchers \cite{chancellor2016quantifying}.
We acknowledge that suicide risk exists on a diverse spectrum~\citep{bryan2006advances}, and a binary distinction is a task simplification intended to alert the human in the loop about exceeding a possible intervention threshold. We note that the studied data is limited to English-speaking Twitter, and also recognize that the data may be susceptible to other demographic, annotator, and medium-specific biases~\citep{hovy2016social}. 
Although our work attempts to analyze aspects of users’ nuanced and complex experiences, we acknowledge the limitations and potential misrepresentations that can occur when researchers analyze social media data, particularly data from a group to which the researchers do not explicitly belong. 

We acknowledge that it is almost impossible to prevent abuse of released technology even when developed with good intentions~\citep{hovy2016social}.
Hence, we ensure that this analysis is shared only selectively and subject to IRB approval~\citep{zimmer2009web} 
 to avoid misuse such as Samaritan’s Radar~\citep{Hsin2016}.
Moreover, we aim to strive for an informed public, by addressing the dual-use threat with preemptive disclosure accompanying the code
, in line with \citet{solaiman2019release}.
Our work does not make any diagnostic claims related to suicide. Our models and analysis should form part of a distributed human-in-the-loop
system for finer interpretation of risk.

\bibliography{anthology,aaai21}

\newpage
\appendix

\section{Differentially Private Optimization}

We implement differential privacy based on the DP-SGD algorithm introduced by \citet{abadi2016deep}. The aim is to control the influence of each of the training data samples during the optimization and hence reduce memorization of sensitive data by controlling the effect of gradients during learning. The key aspects of differentially private SGD are (1) clipping the per-batch gradient where its norm exceeds the clipping bound $C$ and (2) adding Gaussian noise $\mathcal{N}$ characterized by noise scale $\sigma$ to the aggregated per-sample gradients. Given a computed per-sample gradient $\mathbf{g}_i$ belonging to a group of tunable size $N$, the new aggregated gradient $\mathbf{\widetilde{g}}$ is computed by two steps: \\
The gradient clipping step, $G_C(\mathbf{g_i})$ is given by:
\begin{equation}
    \mathbf{\overline{g}}_i \leftarrow \mathbf{g}_i / \max \left( 1,\frac{{\| \mathbf{g}_i \|}_2 }{C} \right)
    \label{eqn:clip}
\end{equation}
The noise addition step, $N_{\sigma}(\mathbf{\overline{g}_i})$ is given by:
\begin{equation}
    \mathbf{\widetilde{g}} \leftarrow \frac{1}{N} \biggl(\sum_{i}\mathbf{\overline{g}}_i + \mathcal{N}(0,\sigma^2C^2\mathbf{I}) \biggr)
    \label{eqn:noise}
\end{equation}

We use a differentially private Adam optimizer, a variant of DP-SGD. This optimization procedure leverages gradient clipping $G_C$ (Equation~\ref{eqn:clip}) and noise addition $N_{\sigma}$ (Equation~\ref{eqn:noise}) to calculate the update $\mathbf{\Delta{w}}$ in the loss landscape as:
\begin{equation}
    \mathbf{\Delta{w}} = -\eta \cdot N_{\sigma}(G_C(\nabla\mathcal{L}(\mathbf{\hat{y}}_i, y_i)))
\end{equation}
where $\eta$ is the learning rate, $\mathcal{L}$ the loss function, and $C$ is the clipping bound.

\section{Impact of Temporal User History and Privacy}:

\label{subsec:Lookback}
\begin{figure}[h!]
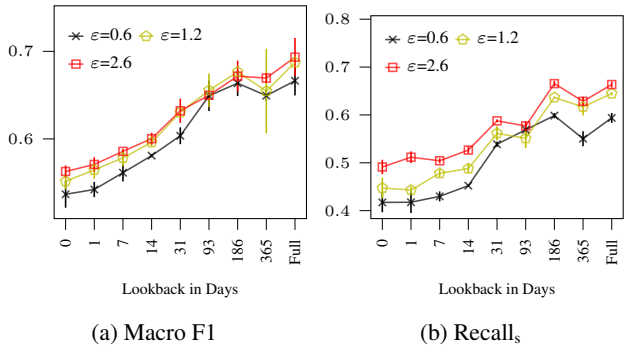

    \begin{subfigure}{0.49\linewidth}
    \includegraphics[width=\linewidth, height=4cm]{plots/lookbackF1Suicide.tikz}
    \caption{Macro F1}
    \label{fig:suicidef1}
    \end{subfigure}
    \begin{subfigure}{0.49\linewidth}
    \includegraphics[width=\linewidth, height=4cm]{plots/lookbackRecallSuicide.tikz}
    \hspace{-8mm} 
    \caption{Recall\textsubscript{s}}
    \label{fig:suiciderecall}
    \end{subfigure}
    \caption{Changes in performance metrics with increasing temporal window on the Suicide Ideation dataset over 10 different runs.}
        \label{fig:lookbackSuicide}
\end{figure}

\section{Experimental Setup}

\noindent \textbf{\underline{Baselines:}}
We discuss the baselines in more detail below.
\noindent \textbf{CurrentPostRF}~\citep{sawhney2018computational}: A non-contextual single-post approach which feeds features such as statistical, LIWC~\citep{pennebaker2001linguistic}, n-grams, and POS counts from the post to a Random Forest (RF) classifier. 

\noindent \textbf{CurrentPostLSTM}~\citep{coppersmith2018natural}: A non-contextual single-post approach which utilizes GloVe embeddings of the post to be assessed fed sequentially to an LSTM.

\noindent\textbf{HistCNN}~\citep{gaur2019knowledge}: A non-sequential model which utilizes GloVe embeddings of posts as a bag of posts which are concatenated and fed to a Contextual Convolutional Neural Network~\citep{shin2018contextual}.

\noindent\textbf{HistDecay}~\citep{mathur2020utilizing}: A sequential model which weighs each GloVe embeddings of historical posts through an exponential decay function and ensembles it with GloVe embeddings trained on a BiLSTM + Attention for the post to be assessed.

\subsection{Tasks and Data Studied}
We use existing tasks and datasets for all experiments.
\label{subsec:datasets}
\paragraph{Suicide Ideation (SI)} We use an existing Twitter dataset curated by \citet{mishra-etal-2019-snap}. These tweets were identified using a lexicon of 143 suicidal phrases, such as ``hate life" or ``want to die". The tweets were annotated by two students of Clinical Psychology, under the supervision of a professional clinical psychologist, achieving a Cohen's Kappa score \citep{cantor1996sample} of 0.72. The annotation guidelines are summarized as follows: \\
\noindent \textbf{Suicidal Intent (SI) Present}: Tweets where suicide ideation or attempts are discussed in a somber, non-flippant tone. \\
\noindent \textbf{Suicidal Intent (SI) Absent}: Tweets with no evidence for risk of suicide, including song lyrics, condolence messages, awareness, news. \\

\paragraph{Suicide Risk (SR)} A user-level multi-class classification task, for which we use the dataset released by \cite{gaur2019knowledge}, which contains 9,127 Reddit posts of 500 users filtered from an intial set of 270,000 users across 9 mental health and suicide related subreddits. The posts were labelled by practicing psychiatrists into 5 increasing risk levels based on the Columbia Suicide Severity Risk Scale, leading to an acceptable average pairwise agreement of 0.79 and groupwise agreement of 0.73. The class distribution of each category with increasing risk level is: Supportive (20\%), Indicator (20\%), Ideation(34\%), Behaviour (15\%), Attempt (9\%). On average, the number of posts made by a user is 18.25$\pm$27.45 with a maximum of 292 posts. The average number of tokens in each post is 73.4$\pm$97.7.

\end{document}